\documentclass[conference]{IEEEtran}
\IEEEoverridecommandlockouts
\usepackage{cite}
\usepackage{amsmath,amssymb,amsfonts}
\usepackage{graphicx, color}
\usepackage{graphicx,subfigure}
\usepackage{subcaption}
\usepackage[export]{adjustbox}
\usepackage{textcomp}
\usepackage{url} 
\usepackage[table]{xcolor} % Needed for cell coloring
\usepackage{paralist}
\usepackage{enumitem}
\usepackage{multirow}
\usepackage{hyperref}
\usepackage{algorithmic}
\usepackage{svg}
\usepackage[linesnumbered,ruled,vlined]{algorithm2e}
\hypersetup{
    colorlinks=true,
    linkcolor=red, % or any color you prefer
    citecolor=blue, % or any color you prefer
    urlcolor=magenta    % or any color you prefer
}
\def\ie{{\em i.e.}}

\def\BibTeX{{\rm B\kern-.05em{\sc i\kern-.025em b}\kern-.08em
    T\kern-.1667em\lower.7ex\hbox{E}\kern-.125emX}}

\newcommand{\smallurl}[1]{\footnotesize\url{#1}}

\definecolor{baselinecolor}{gray}{.9}

\begin{document}

\title{Real-time Lane-wise Traffic Monitoring  in Optimal ROIs}

\author{%
  \IEEEauthorblockN{%
  Mei Qiu, Wei Lin, Lauren Ann Christopher, Stanley Chien$^*$\thanks{$^*$Corresponding Author}, Yaobin Chen, Shu Hu}%
{Purdue University Indianapolis, IN, USA}
}

\maketitle

\begin{abstract}
In the US, thousands of Pan, Tilt, and Zoom (PTZ) traffic cameras monitor highway conditions. There is a great interest in using these highway cameras to gather valuable road traffic data to support traffic analysis and decision-making for highway safety and efficient traffic management. However, there are too many cameras for a few human traffic operators to effectively monitor, so a fully automated solution is desired. This paper introduces a novel system that learns the locations of highway lanes and traffic directions from these camera feeds automatically. It collects real-time, lane-specific traffic data continuously, even adjusting for changes in camera angle or zoom. This facilitates efficient traffic analysis, decision-making, and improved highway safety.
% Our results demo can be found here: \href{https://rb.gy/j86wwh}{\color{blue}Video Demos}.

\end{abstract}

\section{Introduction}
Numerous \textbf{P}an, \textbf{T}ilt, and \textbf{Z}oom (PTZ) traffic cameras are installed along highways in the USA, allowing operators to monitor traffic conditions. However, human operators oversee hundreds of cameras, making it impossible to watch them all simultaneously. Thus, there is interest in using Artificial Intelligence (AI) to analyze footage in real time, providing valuable traffic data and alerting operators to issues. 

Traffic monitoring systems come in various types, broadly classified into two groups: traditional systems and intelligent management systems \cite{nadeem2004trafficview, jain2019review}.
AI and big data analytics have revolutionized traffic monitoring and management. Intelligent Transportation Systems (ITS), equipped with advanced sensors, radars, and license plate recognition cameras play a vital role in detecting and deterring traffic rule violations \cite{abbasi2021deep, bisio2022systematic}. These systems also aid in alleviating traffic congestion by managing traffic scenarios through real-time data visualization \cite{mandhare2018intelligent, mandal2020artificial}. 
% Multiple object detection (MOD) and Multiple object tracking (MOT) are two major technologies used in traffic monitoring system. In MOD, YOLO and FasterRCNN are wildly used in ITS and achieved satisfactory results. 

% Vehicle detection and tracking are two major technologies used in traffic monitoring system. Transfer learning and attention mechanism are frequently used to enhance vehicle detection accuracy under challenging ITS situations \cite{qiu2022attention}, such as low visibility at nighttime, camera blur caused by foggy/rainy/snowy weather, and unsatisfactory camera angles. 
% Some popular multiple object tracking frameworks such as Deepsort \cite{wojke2017simple} FairMOT \cite{zhang2021fairmot} and JDE \cite{wang2020towards} are used in the current traffic surveillance applications \cite{azimjonov2021real, tran2022robust}. 
% \cite{shi2021anomalous}

% MROI
%  The weakness: annotate images from a large video surveillance data, only can monitor two directions' traffic status and not lane-wise designed. existing lane-wise methods : but rely on lane markings; need camera calibration? 

% However, some factors still affect the detection accuracy \cite{zhang2022monocular} such as a lack of automatically detected appropriate region of interest (ROI) in traffic videos.
% ROI is beneficial for excluding noise and vehicles too small to detect for sure.
However, these existing traffic monitoring systems need to annotate images from a large video surveillance data. Most of them can monitor two directions of traffic status but are not lane-specific. There is some work with lane-wise counting, but others rely on road lane marking detection, which fails in various lighting, weather, and ground conditions \cite{zhang2022monocular}. Vehicle motion trajectories can be used to learn lanes \cite{amoguis2023road} but the performance depends on vehicle detection and long tracking accuracy.  The innovation in this work is to perform detection and tracking only in optimal regions of the road. Our new result also solves the problem of existing traffic flow estimation that requires fixed zoom and viewing angles \cite{fedorov2019traffic,zhang2022monocular,li2020trajectory}. 

% \vspace{-4mm}
\begin{figure}
% \vspace{-4mm}
    \centering
{\includegraphics[width=0.48\textwidth]{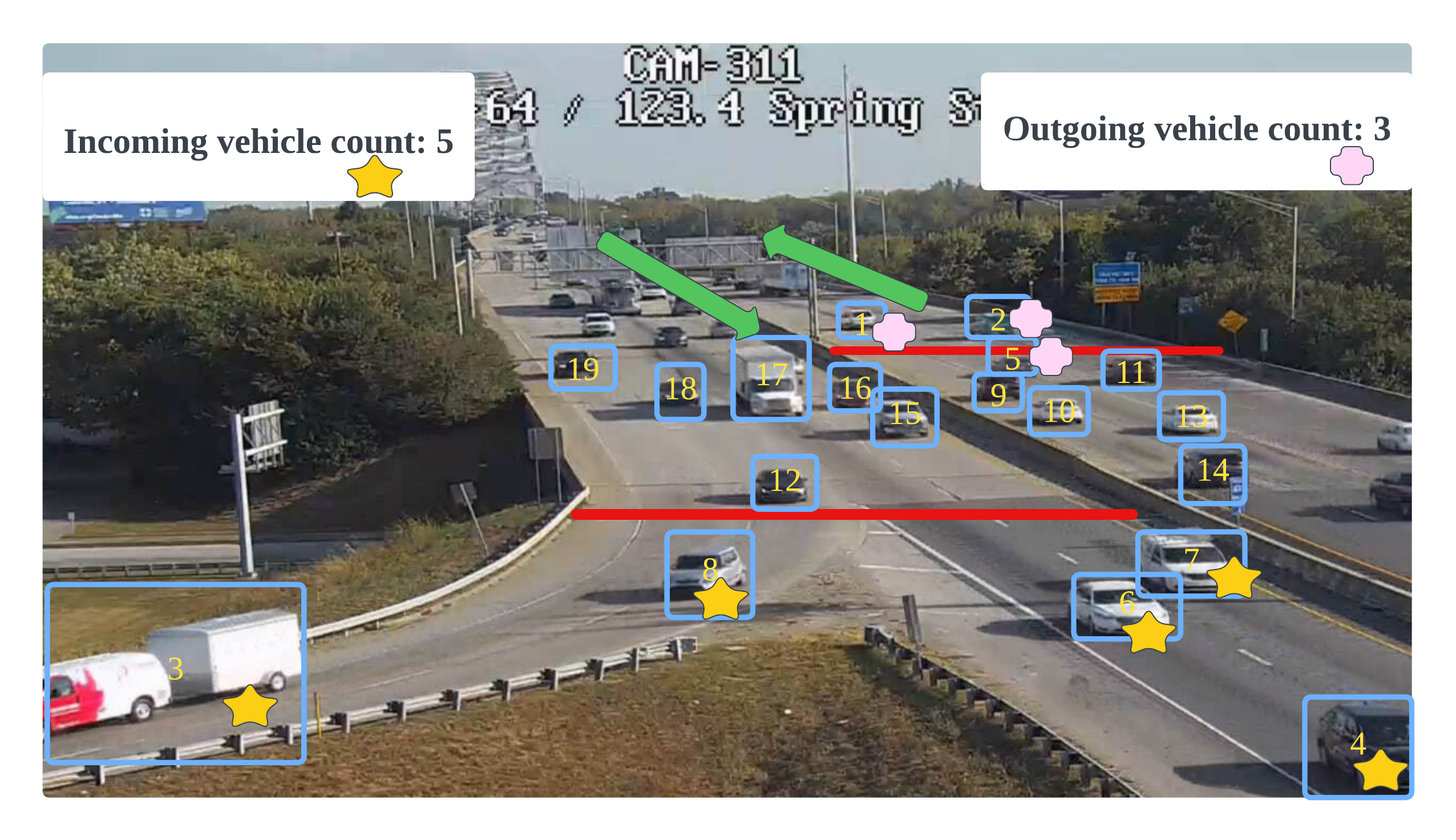}} 
{\includegraphics[width=0.48\textwidth]{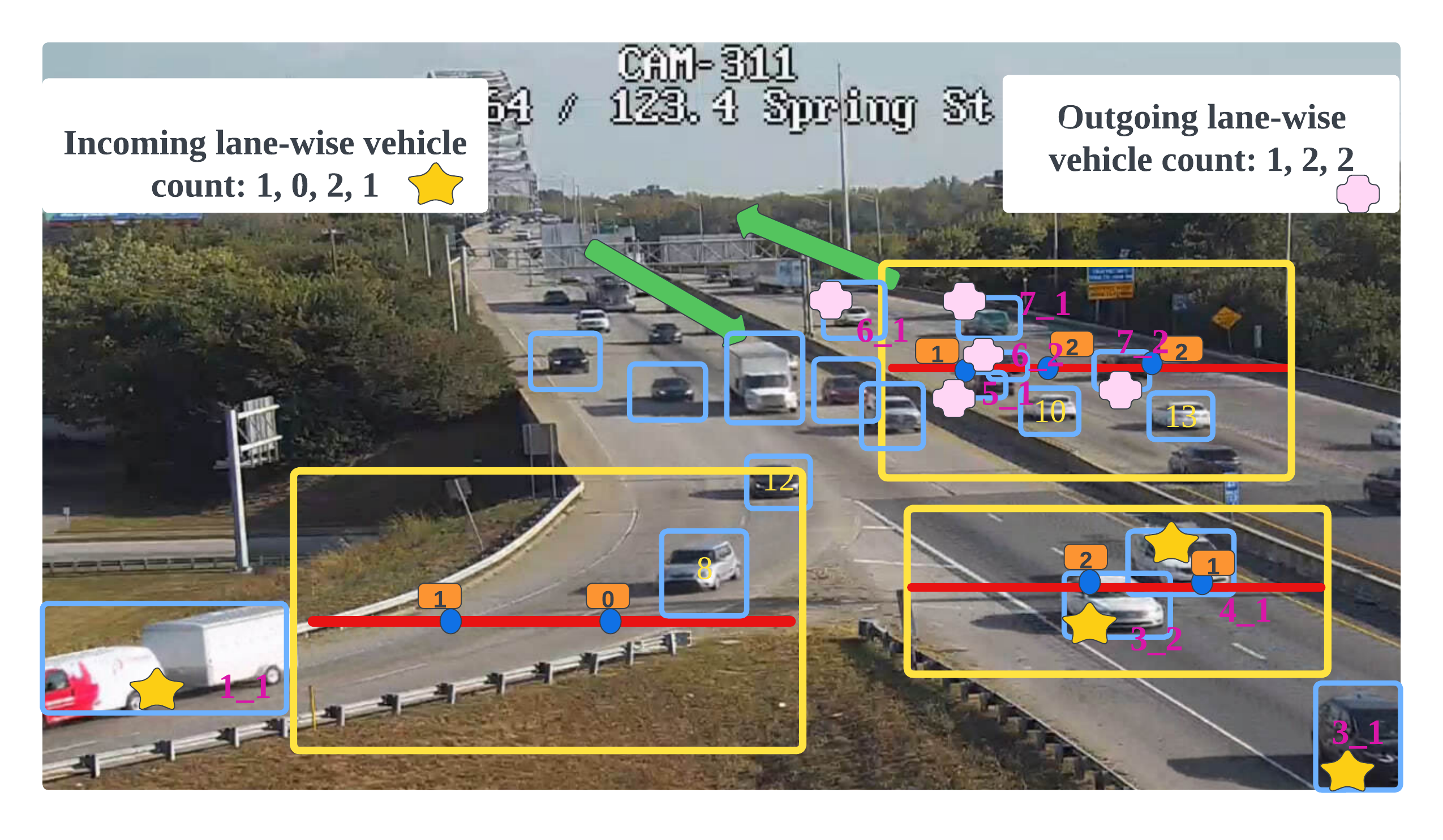}} 
    \vspace{-5mm}
    \caption{\textit{ (Top) \textbf{Traditional System:} Vehicles are detected, tracked, and counted across the entire frame, with manually defined counting lines, leading to suboptimal performance. (Bottom) \textbf{Our system}: Vehicle detection is performed across the entire frame. Vehicle tracking is concentrated in several automatically learned Regions of Interest (ROIs), optimizing detection and tracking results. Counting is lane-specific, with each vehicle assigned a unique LaneID, moving away from bi-directional counting. 
    \textbf{\color{red}{Best view in color}.}}}
    \label{fig:system}
    \vspace{-5mm}
\end{figure}

This paper introduces a Real-Time Automatic Lane Learning and Lane-Specific Traffic Status Monitoring System, building upon our prior research \cite{qiu2021intelligent, qiu2024intelligent}. In our previous work, we developed a method for automatically identifying highway locations, lane boundaries, and traffic flow directions from video footage captured by cameras. Leveraging this acquired knowledge, the current study focuses on the real-time collection and reporting of traffic data for each lane, including metrics such as vehicle counts (illustrated in Fig. \ref{fig:system}), flow rates, and congestion estimates derived from vehicle count statistics. Furthermore, this system is equipped to autonomously detect changes in the camera's angle or zoom level, subsequently reinitiating the road and lane learning process to maintain continuous and accurate traffic status monitoring. 
% The learned road and lane information and the traffic status information are stored in a cloud database. A program on the cloud web server reads and processes the traffic information from the database and brings the results to the user web page. 
% The overall system is depicted in \textbf{Fig \ref{figall}}. 

The contributions of our work can be summarized as follows:
\begin{itemize}
\item Building on our earlier contributions in LCD\cite{qiu2021intelligent} and MRLL\cite{qiu2024intelligent}, this paper introduces an advanced, adaptive system capable of \textbf{lane-wise} vehicle counting, flow rate calculation, and traffic status detection that operates continuously, around the clock. This system is designed to accommodate cameras of varying resolutions and frame rates and is robust against a wide range of weather and traffic conditions, utilizing real-time video streams as its primary input source.
% \item We've developed a new transfer learning approach that combines supervised learning with iterative pseudo-labeling to enhance vehicle detection via YOLOv4, without a significant time investment in labeling. 
\item We have developed a novel, standalone module named ``Camera View Checking" that operates continuously to monitor for any changes in the camera's angle or view. It accomplishes this by comparing the current camera view with an initial, well-defined reference view. Should any deviation in the camera angle or view be detected, this module automatically triggers the system to re-initiate the learning process for road and lane parameters.
\item We devised a novel approach, termed ``Video Rate-Computer Speed Synchronization" to dynamically adjust the input frame rate in accordance with the system's processing capabilities, thereby ensuring the maintenance of real-time performance.
\item We have developed a region-based tracking system leveraging an enhanced DeepSort architecture, specifically designed for real-time tracking in the most effective Regions of Interest (ROIs) to ensure optimal vehicle detection. Key improvements include the integration of our "Video Rate-Computer Speed Synchronization" method, which dynamically adjusts input frame rates based on processing speed to sustain real-time operations. We have also transitioned from using the traditional Intersection Over Union (IOU) metric to the more advanced Complete IOU (CIOU) distance, significantly improving the precision of detection and tracking data association. Furthermore, the introduction of an adaptive matching threshold has been instrumental in optimizing the system's tracking accuracy.
\item We have thoroughly evaluated the effectiveness of our proposed lane-wise vehicle counting system by comparing it with manual counting in 9 videos.
\end{itemize}

\begin{figure*}
\centerline{\includegraphics[width=7.5in]{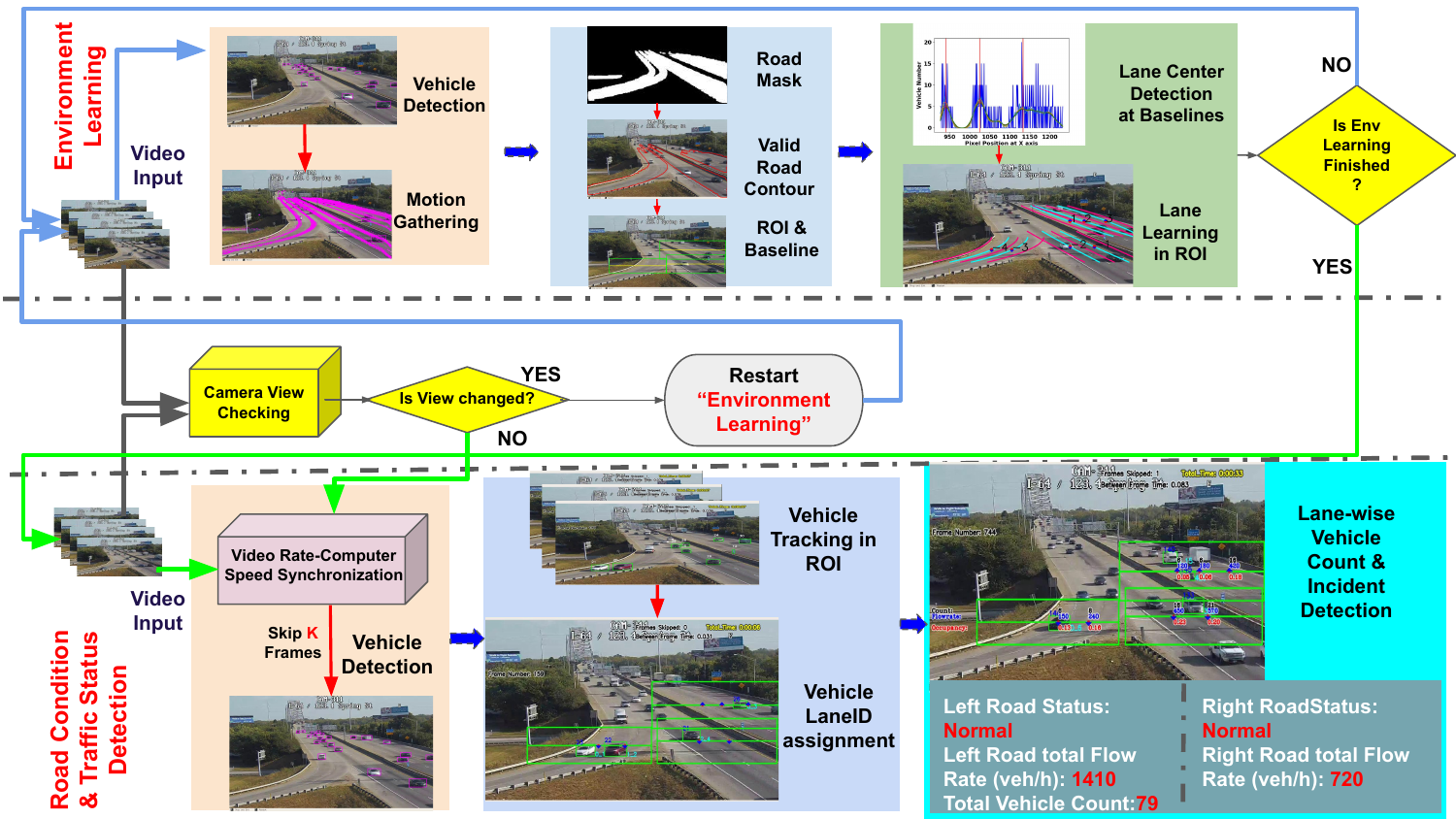}}
\caption{\textit{ The system we designed is running on every local computer with a single GPU. This system gets real-time stream input from a single camera. Since cameras are not fixed on highways, the \textbf{Environment Learning} module will learn the road and lane information based on vehicle motion trajectories (the details can be found in our previous work: \cite{qiu2021intelligent, qiu2022attention, qiu2024intelligent}). When these parameters about road and lanes are learned for a particular view (can be defined by customer), the system will go to the next stage, to do \textbf{Road Condition and Traffic Status Detection} via each lane. An independent module (\ie, \textbf{Camera View Checking}) keeps running all the time to check camera angle changes or not by comparing with the first, well-defined view. Once the system detects the camera angle/view is changed, the Road Condition and Traffic Status Detection module stops working immediately and the system goes to the Environment Learning to learn a new set of road and lane information parameters. More details are explained in Section \ref{sec:method}. 
% The Environment Learning module includes vehicle detection, motion gathering, road mask learning, valid road contour generation, optimal ROIs generation, best baselines in each ROI detection, lane center detection at every baseline, lane curves fitting, lane boundaries, and lane directions detection based on continuous video input. The Road Condition and Incident Detection will do vehicle detection in the whole frame, vehicle tracking only in ROIs, vehicle LaneID assignment in a small region near each baseline, lane-wise vehicle counting, and finally do lane-wise flow rate and traffic status estimation and incident detection. A particular method ``Video Rate-Computer Speed Synchronization" is designed to estimate how many frames need to skip at each frame input, so \textbf{K} is dynamic. This algorithm only works in Road Condition and Incident Detection. {\color{red} Road Condition and Incident Detection only starts working when the environment learning  tasks are all finished and camera view is not changed. In Fig. \ref{figedg}, the bold, solid arrows show the flow of feedback to the input with the signals of whether environment learning is finished and the camera angle changes or not. The dashed blue arrow shows the continuous video input to the environment learning module and the dashed green arrow shows the video input to the Road Condition and Incident Detection module after \textbf{K} frames are skipped. The two modules work in a sequential order.}
}\label{figedg}}
\end{figure*}
% \begin{figure}
% \centerline{\includegraphics[width=3.5in]{images/existing_count.png}}
% \end{figure}
\begin{figure}[t]
        \centering
        \subfigure[]{
            \includegraphics[width=0.45\linewidth]{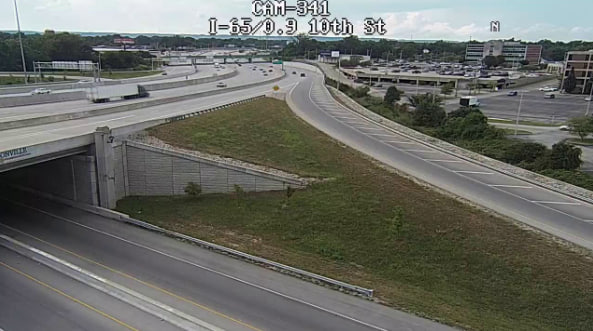}
            \includegraphics[width=0.45\linewidth]{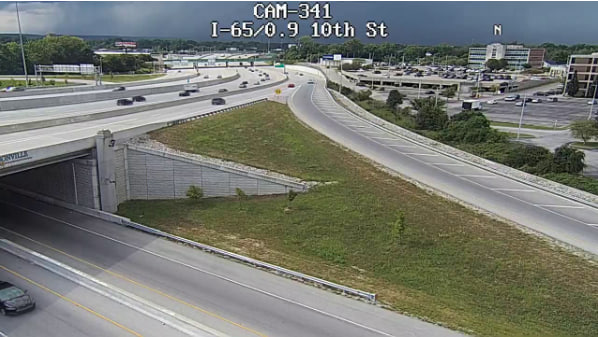}
        }
        \subfigure[]{
            \includegraphics[width=0.45\linewidth]{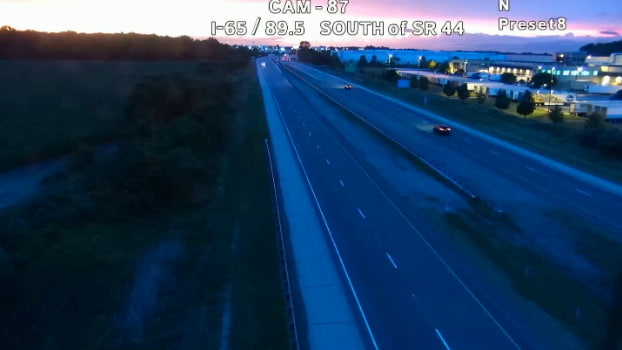}
            \includegraphics[width=0.45\linewidth]{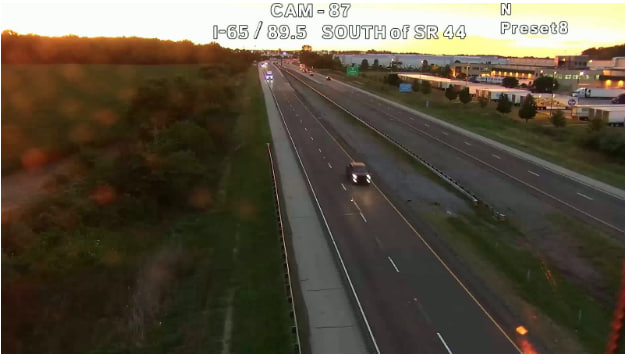}
        }
        
        \caption{\textit{Correct Camera View Change Detection Cases. (a) The cloud background of the right frame changed significantly from the left frame but did not cause false detection. The system did not report the camera view change. This is \textbf{correct detection}. (b) The left frame is taken at dawn and the right frame in the afternoon. The system did not report the camera view changes. This is \textbf{correct detection}.} }
        \label{fig4}
\end{figure}
\begin{figure}[t]
        \centering
        \subfigure[]{
            \includegraphics[width=0.45\linewidth]{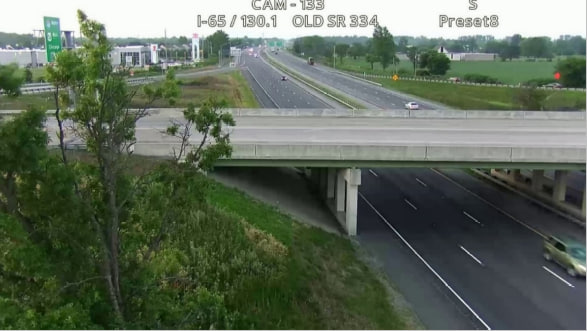}
            \includegraphics[width=0.45\linewidth]{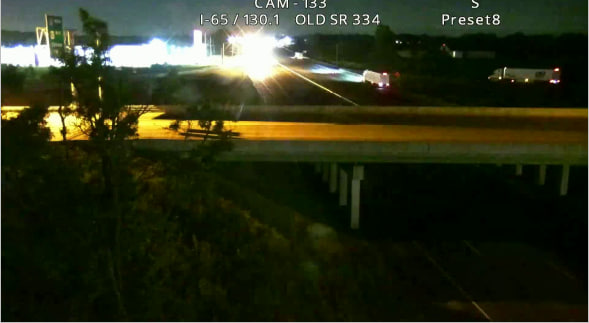}
        }
        \subfigure[]{
            \includegraphics[width=0.45\linewidth]{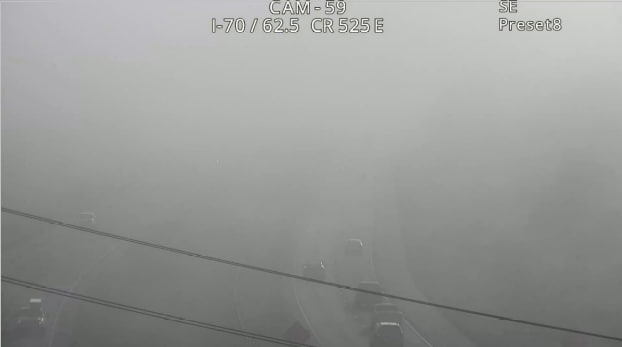}
            \includegraphics[width=0.45\linewidth]{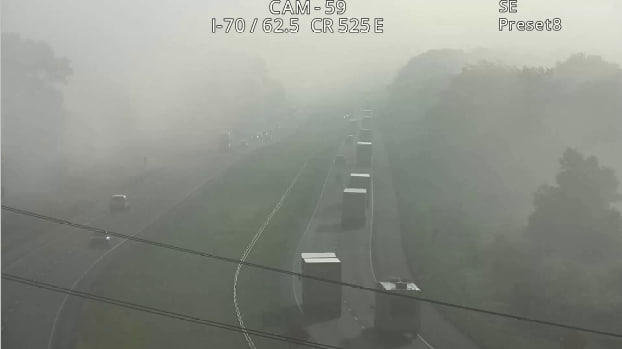}
        }
       
        \caption{\textit{Wrong Camera View Change Detection Cases. (a) The left frame is taken in the morning and the right frame is taken at night. The algorithm gives false detection because of the flare in the second frame is background. The Hash value distance returned by our algorithm is larger than the threshold. This is \textbf{wrong detection}. (b) The algorithm predicts camera angle is changed in the right frame from the left frame while it is not. But it is very tough for humans as well to predict in this kind of scenario. This is \textbf{wrong detection}.}} 
        \label{fig5}
\end{figure}

\section{Related Works}
% Since vehicle detection and tracking are the key components in lane learning and traffic status gathering, object detection and tracking methods were first investigated.
\subsection{Object Detection}
Traditional object detection methods like background subtraction \cite{cai1995tracking} and image difference \cite{dailey2000algorithm} have been replaced by deep learning, due to superior detection accuracy and speed. Deep learning methods fall into two categories: two-stage detectors (e.g., Faster R-CNN, Mask R-CNN \cite{ren2015faster, he2017mask}) and one-stage detectors (e.g., YOLO, YOLOv4 \cite{redmon2016you, bochkovskiy2020yolov4}). Two-stage detectors offer superior accuracy, while one-stage detectors provide faster inference without pre-generated region proposals. These deep learning-based object detection algorithms are extensively employed in traffic surveillance videos, delivering satisfactory performance \cite{shi2021anomalous, mitra2021towards}.
 % However, some factors still affect the detection accuracy \cite{zhang2022monocular} such as a lack of automatically detected appropriate region of interest (ROI) in traffic videos. ROI is beneficial for excluding noise and vehicles too small to detect for sure. However, the vehicle detector’s performance deteriorates in highly challenging environments, such as low visibility at nighttime, camera blur caused by foggy/rainy/snowy weather, unsatisfactory camera angles, and overlapped vehicles in congested traffic.
\subsection{Multiple Object Tracking}
Most object-tracking algorithms fall into two categories: detection-based tracking (tracking-by-detection) and detection-free tracking. In recent years, tracking-by-detection methods have been predominant. These methods involve performing object detection on each frame to obtain detection results, which are then associated with adjacent frames to form trajectories. Popular data association methods include the Hungarian algorithm and the Kuhn-Munkres algorithm \cite{kuhn1955hungarian}. Post-processing techniques like soft non-maximum suppression (NMS) \cite{bodla2017soft} are often applied to smooth and refine the trajectories. The Intersection Over Union (IOU) \cite{bochinski2017high} metric is commonly used to associate detection results based on bounding box overlaps between frames. SORT \cite{bewley2016simple} and Deep SORT \cite{wojke2017simple} are two widely used tracking methods in traffic surveillance videos \cite{liu2021city, yu2018traffic}. SORT combines Kalman Filter with the Hungarian algorithm and relies on bounding box size and position for motion estimation and data association. Deep SORT utilizes a CNN to extract appearance information, enhancing the association metric with motion information. This allows Deep SORT to track objects more effectively during occlusions, reducing ID switches but also increasing computational costs.
% Other popular multiple object tracking frameworks such as FairMOT \cite{zhang2021fairmot} and JDE \cite{wang2020towards} are used in the current traffic surveillance applications \cite{somaldo2021comparison, tran2022robust}. The limitation of the tracking-by-detection framework is that its performance is highly dependent on vehicle detection performance, which should be considered carefully.

\section{Method}\label{sec:method}
This section summarizes our system, described in \cite{qiu2021intelligent, qiu2024intelligent}, that analyzes highway surveillance camera footage in real-time to detect changes in camera angles or zoom levels. It then learns the number of lanes and their locations, providing lane-wise vehicle counting, flow rate, and congestion estimation in optimal ROIs with best vehicle detection. The system, depicted in \textbf{\textbf{Fig. \ref{figedg}}}, runs on a local computer with a single GPU and comprises two main components: \textbf{Environment Learning} for adapting to unpredictable camera settings and \textbf{Road Condition and Traffic Status Detection} for analyzing lane-based traffic information. 
\subsection{Environment Learning}
The goal of this component includes finding the locations of the lanes and optimal locations to detect and count vehicles on the lanes. Instead of relying on non-robust lane-markings in highways, we adopted a method that uses the vehicle motion information on the video to find the lane locations because most vehicles on the highway are driven within their lanes \cite{qiu2021intelligent, qiu2024intelligent}. 

\textbf{Camera View Checking}
The program monitors for changes in camera angle or zoom level which would invalidate learned lane information. The efficient ImageHash algorithm \cite{marr1980theory} is used to assess significant and rapid background changes in the top fifth of the image frame (the sky and far side of the road), to signal potential camera adjustments. If the Hash Distance value exceeds a certain threshold, it is considered a camera angle or zoom level change, prompting a relearning process. This check occurs every 50 frames.

The algorithm is proven to be very robust in various camera motions and lighting conditions. The example testing cases are shown in Fig. \ref{fig4}, such as when the cloud background changed significantly, when the two input videos were taken over a long time apart, or when the camera zoomed out or zoom in, under all of these tough PTZ camera situations, our algorithm gives the correct detection with the default Hash threshold. This algorithm detects the large camera angle shifts and also detects those less apparent camera parameter changes. There are some cases that our algorithm cannot handle well, as shown in Fig. \ref{fig5}.
In summary, our camera viewing checking algorithm works well in the majority of our ITS environments. The system monitors the camera's viewing angle. If it detects changes in angle or zoom level, it halts and restarts the environment learning stage to relearn the road and lane parameters from the video.

\subsection{Road Condition and Traffic Status Detection}
Learning the ROI and lane boundaries makes lane-based traffic status detection possible. The issues faced and solutions in lane-based traffic status detection are described in the following subsections.
\subsubsection{Video Rate-Computer Speed Synchronization}
The goal of this system development is that it can be used for all highway surveillance cameras. Since the frame rate (frame per second: $fps$) and resolution of these cameras can be quite different, the processing time for image processing can vary significantly. We reduce the $fps$ processed based on the computer computation speed(assume the time cost of each frame is $\delta t$ with the unit of second) by adaptively skipping $K$ frames for every frame processed during the process following {Equation (\ref{formula-3})}:
 \begin{equation}
\label{formula-3}
 K = 
 \begin{cases}
    fps - \frac{1}{\delta t} &\text{if $\frac{1}{\delta t}-fps<0$} \\
    0 & \text{otherwise}
  \end{cases}
\end{equation}
We designed a Video Rate-Computer Speed Synchronization method that observes the computer processing time for each input frame and the incoming frames per second. Then we use this information to determine how many frames should be skipped for each frame processed so there will be no backlogged frames. Since the processing time for each frame is affected by the vehicles captured and processed in the frame, the number of frames skipped fluctuates over time.

\subsubsection{Vehicle Detection}
The pre-trained YOLOv4 is used in this work and  the input to vehicle detection is no longer continuous but with $K$ frames skipped.
\subsubsection{Online Multi-target Tracking with Adaptive Kalman Filter}
Before tracking, we filter out invalid detected vehicles, removing those with bounding box sizes more than twice the median truck size and those outside any ROI or roads. NMS is applied to eliminate redundant bounding box overlaps. DeepSort \cite{wojke2017simple} is then employed for online multiple vehicle tracking on the remaining valid detection in each frame.
However, the classic tracking pipeline has limitations in our application. Firstly, real-time performance on a single GPU is essential, but feature extraction from a deep learning model at each frame is time-consuming. Secondly, our proposed skipping frame strategy renders the IOU distance and fixed threshold in DeepSort inadequate, as detection and tracklets may not overlap. Therefore, we modify DeepSort to meet our application's requirements.

\textbf{We modified Deepsort:}
\textbf{By removing object appearance feature matching and using CIOU distance instead of IOU distance in Cascade Matching step in DeepSort}\\
By importing the frame-skipping strategy, the whole system achieves real-time processing status without much tracking accuracy loss. However, in the default, Deepsort, the IOU distance used in the last step for matching detected vehicles and tracks will not meet the requirement if the new detected vehicles do not overlap. So the distance between the bounding box's centers and the consistency of the aspect ratio has the potential to solve this problem. Based on the work CIOU loss \cite{zheng2021enhancing}, we replace the IOU distance with CIOU distance. The original IOU distance is defined as Equation (\ref{eqn:diou}) and IOU is the intersection-over-union of bounding boxes in detections and tracks. The modified CIOU distance is defined as shown in Equation (\ref{eqn:ciou}). 
\begin{equation}
\label{eqn:diou}
d_{IOU}=1-IOU
\end{equation}
\begin{equation}
\label{eqn:ciou}
d_{CIOU}=1-IOU+ D + \alpha*V
\end{equation}
In CIOU, the normalized
central point distance $D$ is designed to measure the distance of two boxes as calculated in Equation (\ref{eqn:D}), 
\begin{equation}
\label{eqn:D}
D = \frac{\rho^{2}(p^{d}, p^{t})}{c^{2}}
\end{equation}
where $p^{d} = [x_{d}, y_{d}]^{T}$
and $p^{t} = [x_{t}, y_{t}]^{T}$ are the central points
of boxes in detections and tracks
, $c$ is the diagonal length of each box in the track, and
$\rho$ is specified as the Euclidean distance function. $V$ is defined as the consistency of the aspect ratio and calculated as: 
$\frac{4}{pi^{2}}(\arctan \frac{w^{d}}{h^{d}} -\arctan \frac{w^{t}}{h^{t}})$. The trade-off parameter $\alpha$ is defined the same as with the default CIOU. 
The CIOU distance mitigates the missing match when detections are not overlapping with predicted tracks and are more robust to the scale of bounding boxes.

\textbf{The matching distance threshold was adjusted with the skipped frame number $K$ in Cascade Matching step in DeepSort}\\
Based on the above analysis, a fixed IOU distance threshold does not work well because of different frames are skipped in each iteration. So we adjust the CIOU distance threshold to $pre_{thre}*(K+1)$, where $K$ donates the skipped frame number and $pre_{thre}$ is the default fixed matching threshold in Deepsort, if no frame is skipped, $K$ equals 0. 

After tracking, each vehicle passing through the ROI is assigned a unique $Fid$.
\subsubsection{Lane-wise Vehicle Counting}
The result of vehicle counting with time information can be proportionally converted to flow rate and valuable traffic status information. To realize lane-wise vehicle counting, we associate every tracked vehicle $i$ in the ROI with $Fid_{i}$ to a unique lane ID $Lid_{i}$ using the lane boundary information obtained in the environment learning stage. 
When assigning the lane ids to each vehicle in the current frame, only the previous frame is compared. This strategy will not only reduce the time cost and also reduce the id switch when we assign the lane ids if the id switch happens in FIDs in a longer trace. If the center of a tracked vehicle is within the boundary of a lane and within the average car length at the baseline, the vehicle is counted for that lane. A tracked vehicle is only counted one time. 
\subsubsection{Traffic Status Detection}
The system can generate traffic status and incident reports for the users in three steps: (i) determine the flowrate of each lane based on the vehicle counts over a specific time interval periodically (ii) estimate the percentage of pixels occupied by all vehicles on each lane within the ROI as occupancy rate as shown in the \textbf{Equation (\ref{formula-5})}, and (iii) traffic status checking based on the combination of the flow rate and occupancy.
\begin{equation}
\label{formula-5}
 Occp_{l} = \frac{\sum(h_{l})}{H}
\end{equation}, here $l$ is the lane ID, and $h_{l}$ is the sum of all vehicles' bounding box height in lane $l$.  
For every $T$ minute, the instantaneous flow rate $Fr_{l}$(vehicle per hour) of lane $l$, is calculated based on the vehicle counting $C_{l}$ of lane $l$ during $T$: $Fr_{l} = \frac{C_{l}*60}{T}$. The time interval for the flow rate calculation cannot be too long since it will not show the short-term traffic status changes. On the other hand, the time interval for the flow rate calculation cannot be too short since it will fluctuate too much to understand its meaning.

 \begin{figure}
\centerline{\includegraphics[width=1\linewidth]{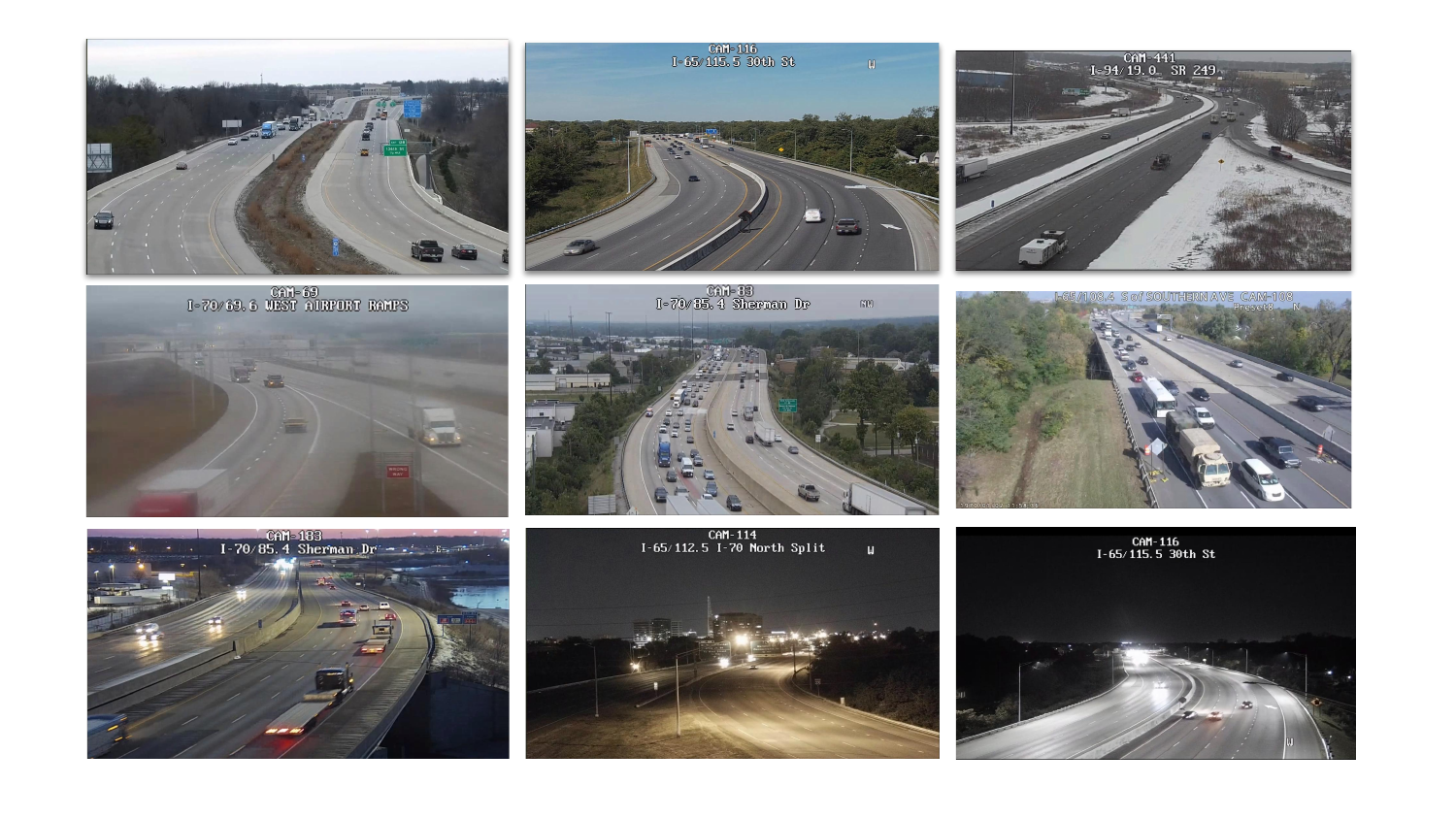}}
% \centerline{\includegraphics[width=3.5in]{images/ourcount.svg}}
\vspace{-3mm}
\caption{\textit{9 ITS traffic scenes recorded in sunny, rainy, snowy, nighttime, and congestion traffic conditions.}}
\label{fig:data}
\end{figure}

 % \begin{figure*}
% \centerline{\includegraphics[width=7in]{images/webserver.pdf}}
% \caption{System operation with multiple cameras and user interface. A zoomable map with all installed cameras is provided on the top left to allow the user to select the camera and location of interest. After a camera is selected on either the map or the list, it is marked on the map. Then the most recent flow rate of lanes seen by this camera is presented on the bottom left. Whether a camera video is being learned or already learned is shown in the middle bottom. The table on the right of the window provides a list of cameras. The list is sorted according to their importance for the operator to give attention—for example, cameras seeing jam traffic rank higher than cameras seeing slow traffic. The data download feature allows users to gather lane-specific data for specific cameras and time intervals.
% \label{figweb}}
% \end{figure*}

% \subsection{User Interface Design}
% This system is developed with two primary purposes: to provide alerts to the traffic management operator for situations of concern and to keep the lane-specific traffic data for long-term traffic pattern analysis. 

\section{Experiments}
\subsection{Experimental Settings}
\smallskip
\noindent
\textbf{Datasets.}
For vehicle count testing purpose, we created another 9 videos and each video lasts about 2 minutes. These testing videos have unique camera views and various weather, traffic density and visibility as shown in Fig. \ref{fig:data}. We assume each video has finished the ROIs and lane learning including: lane center locations, lane directions, lane boundaries in all the ROIs.

\smallskip
\noindent
\textbf{Evaluation Metrics.}
To estimate our system's performance in vehicle counting in each lane, we manually counted 9 videos from four scenes including sunny, rainy, night, and congested traffic, and each video lasts 2 minutes. We counted vehicles in each lane separately when every vehicle passes the baseline. The flowrate of ground truth is also estimated with the counting ground truth. The total counting accuracy of one video is defined as the corrected counting percentage: $\frac{total\quad system\quad count}{total\quad ground \quad truth\quad count}*\%$. We consider the smaller count as the numerator and the higher the value means the higher the counting accuracy.
We design two other matrices to estimate the road level flow rate accuracy compared with the flow rate generated from the ground truth counting: \textbf{MEA}(mean estimation accuracy): $\frac{\sum(Fr_{l})}{\sum(FrGT_{l})}$ and \textbf{RMSE}(root mean square error): $\sqrt{\frac{\sum{(Fr_{l}-FrGT_{l})^2}}{m}}$. We define the $FrGT_{l}$ as the ground truth flow rate at lane $l$ and $m$ as the total number of  lanes. The MAE has a similar definition of counting accuracy, we regard the smaller value as the numerator and the higher value means the estimation is closer to the ground truth. The RMSE estimates the whole error of estimation, and the lower value means better counting and flow rate estimation.
To check the traffic status, we defined an estimate rule based on the flow rate and  occupancy rate of each lane as shown in Equation (\ref{formula-7}).
Based on the rules we defined, our system gets 100\% accurate traffic status (normal, slow, and jam) reports of all the lanes of total 9 videos.
\begin{equation}
\label{formula-7}
 status\_{l} = 
 \begin{cases}
    Jam & 
    \substack{\text{if $Fr_{l}<600$ and $Occp_{l}>0.6$} }\\
    Slow & \substack{\text{if $600<Fr_{l}<900$ and $0.4<Occp_{l}<0.6$}}
    \\
    Normal & \text{otherwise}
  \end{cases}
\end{equation}
\smallskip
\noindent
\textbf{Manual Count.}
We hire 2 master students to count the actual vehicles crossing the counting line manually. For each video, they count the vehicles in each video on a regular interval which is 30 seconds and totally count 4 intervals. And we also record flow rate value, and occupancy in each lane during the same interval calculated automatically by our designed system. Finally, we analyze how these two indicators change over time and do the comparison.

\smallskip
\noindent
\textbf{Implementation Details.}
We utilized YOLOv4 \cite{bochkovskiy2020yolov4}, a popular deep learning object detector, to detect vehicles in camera frame streams. During the reference phase, we set detection confidence score thresholds to 0.25 and IOU threshold to 0.45. For tracking, we employed Deep Sort framework, excluding the CNN feature extractor. Cosine distance metric facilitated track association in each frame, with an initial IOU distance threshold of 0.35. Parameters like $max\_age$ (30) and $n\_init$ (3) were set to control track deletion and initialization phases, respectively. Coupled with our frame-skipping strategy, the entire system ran in real-time on a single NVIDIA Quadro RTX 5000 GPU.

% \begin{figure}
% \centerline{\includegraphics[width=3.5in]{images/edge_processing.png}}
% \caption{An }
% \label{fig:resu}
% \end{figure}

\subsection{Results}

\begin{figure}
\centerline{\includegraphics[width=1\linewidth]{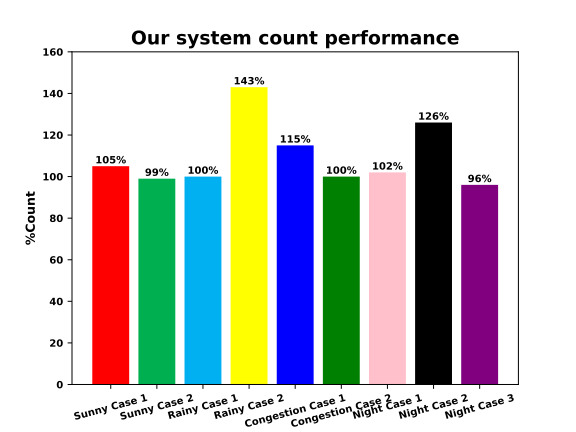}}

% \includegraphics[width=1\linewidth]{images/ourcount.png}
% \centerline{\includegraphics[width=3.5in]{images/ourcount.svg}}
\vspace{-3mm}
\caption{\textit{The overall count percentage for all vehicles from 9 cases by our counting system. Lower than 100\% means miss count, larger than 100\% means over count, and 100\% means count by our system is totally the same as the ground truth.}}
\label{figcount}
\vspace{-6mm}
\end{figure}

\noindent
\textbf{Lane-wise Real-time Vehicle Counting.}
The lane detection in this system performs with acceptable accuracy when not in totally dark conditions. Experimental results on our own collected highway data demonstrates the effectiveness of our proposed framework in lane-wise vehicle counting by comparing with manual ground truth. The system count of 7 of 9 videos is close to the ground truth, and lowest counting error is 2\% as shown in Fig.\ref{figcount}.
However, the performance decreases in low lighting conditions, poor viewing angles, or with occlusions.

\smallskip
\noindent
\textbf{Lane-wise Flow Rate Estimation.}
When estimating our system's lane-wise flow rate estimation accuracy, we only consider the cases in which all the lanes are correctly learned. The flow rate estimation accuracy and error results of the 8 videos are shown in \textbf{TABLE \ref{table:3}}. 

These results prove our system achieves good flow rate estimation accuracy and the highest MAE reaches \textbf{0.95} and it reaches 0.86 even in night cases. The average RMSE is lowest in the sunny and daytime cases which means the lighting is still the major factor affecting the detection and counting.

\smallskip
\noindent
\textbf{Road Condition and Traffic Status Detection.}
Based on the rules we defined, our system gets 100\% accurate traffic status (normal, slow, and jam) reports of all the lanes of all 9 videos.

\begin{table}[t]
\centering
\setlength\tabcolsep{12pt} % default value: 6pt
    \caption{\textit{Accuracy of lane-wise flow rate estimation in various scenarios.}}
  \footnotesize
    \begin{tabular}{*{3}{c}}  
      \hline
      \textbf{Video Name} &\textbf{MEA $\uparrow$} &\textbf{RMSE $\downarrow$} \\
       \hline
       Sunny 1& \textcolor{red}{\textbf{0.95}}& 76.49\\
       \hline
       Sunny 2 &0.93 &54.38\\
       \hline
       Rainy 1 &0.93 &47.43\\
      \hline
       Rainy 2 &0.77 &228.47\\
       \hline
        Night 1 &0.86 &118.59\\
        \hline
         Night 2 &0.93 &\textcolor{red}{\textbf{36.74}}\\
         \hline
          Congestion 1 &0.86 &354.96\\
          \hline
           Congestion 2 &0.9 &129.03\\
           \hline
            Average &0.89&130.63\\
           \hline
\end{tabular}
    \label{table:3}
\end{table}

\section{Conclusion}
Highway surveillance cameras face challenges due to unpredictable changes in viewing direction and zoom level. To address this, a lane-based automatic traffic monitoring system with a lane learning component has been developed and proved its success using real-time Indiana Highway data. The paper outlines the system's architecture and highlights the importance of vehicle motion for lane detection, vehicle counting and traffic status estimation.

\textbf{Limitations.}
The performance of our system depends critically on the accuracy of its various sub-modules, which include lane learning, vehicle detection, tracking, LaneID assignment, and counting. Specifically, lane learning, which is primarily informed by video input and vehicle motion, tends to be less effective in situations such as traffic jams or when vehicles remain stationary. The task of vehicle detection faces significant challenges in the Intelligent Transportation System (ITS) environment, where adverse weather conditions and low-light conditions at night can significantly impair detection capabilities.

In Multi-Object Tracking (MOT), one of the major hurdles is dealing with occlusions, particularly in dense traffic scenarios where bounding boxes of different vehicles overlap, making it difficult to track individual vehicles accurately. Moreover, the use of heavy-weight computational architectures for tracking hampers the ability to perform real-time tracking, necessitating the development of lighter-weight alternatives that do not sacrifice accuracy.

To accurately assess the overall performance of our system, it is essential to conduct tests across a broader range of scenarios to validate the system's robustness. This entails incorporating more diverse test cases that can effectively simulate the wide variety of real-world conditions under which the system must operate.

\textbf{Future Work.}
To improve the overall accuracy of the system, it is crucial to make incremental improvements to each individual module. Additionally, employing a more diverse collection of test videos will be instrumental in accurately estimating the system's performance.

% \section*{Acknowledgment}

\smallskip
\noindent
\textbf{Acknowledgment}.
This work was supported by the Joint Transportation Research Program (JTRP), administered by the Indiana Department of Transportation and Purdue University, Grant SPR-4436. The authors would like to thank all
INDOT Study Advisory Committee members, including Jim Sturdevant, Ed Cox, Tim Wells, for their guidance and advice
throughout the project. The authors would also thank Zhengming Ding, Stephan Gerve, Upadhyay, Aniket Pankaj, Prathamesh Somkant Panat, Anup Atul Mulay, Baudouin Ramsey, Kunal Mandil, Arjun
Narukkanchira Anilkumar, and Kavya Prasad for their contributions in various part of this project.

% \small
% \bibliographystyle{plain}
\bibliography{main}

\end{document}